\documentclass[letterpaper, 10 pt, conference]{ieeeconf}  

\IEEEoverridecommandlockouts                              

\usepackage{graphicx} 
\usepackage{amsmath} 
\usepackage{algorithm}
\usepackage{algorithmic}
\usepackage{array}
\usepackage{eqparbox}
\usepackage{soul}
\usepackage{color}

\usepackage{setspace}
\usepackage{dblfloatfix}
\title{\LARGE \bf
Pedestrian Motion Model Using Non-Parametric Trajectory Clustering and Discrete Transition Points*
}

\author{Yutao Han, Rina Tse, and Mark Campbell
\thanks{*This research is funded by the Office of Naval Research Grant N00014-17-1-2175.}
\thanks{Yutao Han is a PhD student in the Department of Mechanical Engineering, 
		Cornell University, 
        {\tt\small yh675@cornell.edu}.}%
\thanks{Rina Tse is a Postdoctoral research associate in the Department of Mechanical Engineering, 
		Cornell University, 
        {\tt\small rt297@cornell.edu}.}%
\thanks{Mark Campbell is a Professor in the Department of Mechanical Engineering, Cornell University,
        {\tt\small mc288@cornell.edu}.}%
}

\begin{document}

\clearpage
\thispagestyle{empty}
\onecolumn
\noindent \textcopyright 2019 IEEE. Personal use of this material is permitted.  Permission from IEEE must be obtained for all other uses, in any current or future media, including reprinting/republishing this material for advertising or promotional purposes, creating new collective works, for resale or redistribution to servers or lists, or reuse of any copyrighted component of this work in other works.

\bigskip

\noindent This is the author manuscript without publisher editing. An edited version of this manuscript was published in Robotics and Automation Letters (RA-L). To access the official IEEE published version use the identifiers below:

\bigskip

\noindent DOI: 10.1109/LRA.2019.2898464

\bigskip

\noindent URL: https://ieeexplore.ieee.org/document/8638524

\bigskip

\noindent \textbf{Cite as}:

\noindent Y. Han, R. Tse and M. Campbell, "Pedestrian Motion Model Using Non-Parametric Trajectory Clustering and Discrete Transition Points," in IEEE Robotics and Automation Letters, vol. 4, no. 3, pp. 2614-2621, July 2019.
doi: 10.1109/LRA.2019.2898464

\bigskip

\noindent \textbf{BibTex}:

\noindent @ARTICLE\{8638524,

\noindent author=\{Y. \{Han\} and R. \{Tse\} and M. \{Campbell\}\},

\noindent journal=\{IEEE Robotics and Automation Letters\},

\noindent title=\{Pedestrian Motion Model Using Non-Parametric Trajectory Clustering and Discrete Transition Points\},

\noindent year=\{2019\},

\noindent volume=\{4\},

\noindent number=\{3\},

\noindent pages=\{2614-2621\},

\noindent doi=\{10.1109/LRA.2019.2898464\},

\noindent ISSN=\{2377-3774\},

\noindent month=\{July\},\}

\twocolumn
\clearpage

\maketitle
\thispagestyle{empty}
\pagestyle{empty}

\begin{abstract}

This paper presents a pedestrian motion model that includes both low level trajectory patterns, and high level discrete transitions. The inclusion of both levels creates a more general predictive model, allowing for more meaningful prediction and reasoning about pedestrian trajectories, as compared to the current state of the art. The model uses an iterative clustering algorithm with (1) Dirichlet Process Gaussian Processes to cluster trajectories into continuous \textit{motion patterns} and (2) hypothesis testing to identify discrete transitions in the data called \textit{transition points}. The model iteratively splits full trajectories into sub-trajectory clusters based on transition points, where pedestrians make discrete decisions. State transition probabilities are then learned over the transition points and trajectory clusters. The model is for online prediction of motions, and detection of anomalous trajectories. The proposed model is validated on the Duke MTMC dataset to demonstrate identification of low level trajectory clusters and high level transitions, and the ability to predict pedestrian motion and detect anomalies online with high accuracy. 

\end{abstract}

\section{INTRODUCTION}

\subsection{Motivation}

Predictive models allow systems interested in pedestrian behavior to perform higher level inference and reasoning about scenes involving pedestrians. Models of how pedestrians navigate an environment are useful for many robotics applications, including: surveillance robots \cite{witwicki2017autonomous} (e.g. understanding patterns of how people walk in a subway station could help detect anomalies, such as a person running with a bag after a theft) and search and rescue robots \cite{murphy2008search} (e.g. understanding movement patterns in a building on fire or during an earthquake can give information about where people may be in the building). 


Currently there are advanced and accurate systems for tracking multiple targets with cameras within a scene \cite{MultiTrackDynamicBayes}, \cite{MKernelTrack}. However, to gain more insight into the scene and behaviors, patterns, motivations, anomalies etc., a more comprehensive model of pedestrian motions and behaviors is required. In this paper, a non-parametric model is developed to capture the statistics of pedestrian trajectories and reason about pedestrian motion. Given the challenges of physics based modeling of pedestrian motion and the availability of large datasets of pedestrian movements, data-driven models are intuitively appropriate. Data-driven models do not make any assumptions about pedestrian motion, such as optimizing trajectory distance or avoiding collisions with other pedestrians that may constrain the  model to produce unrealistic results. Instead, data-driven models have the flexibility to fit to whatever the data suggests. Machine learning techniques can be used to develop predictive models much smaller in size than original datasets. In this paper, descriptive transition points are learned from the data to represent high level behaviors where trajectory patterns branch or merge together (for example, at a fork in the road). These transition points are learned in conjunction with low level trajectory clusters in an iterative algorithm to form a general model of pedestrian motion. By integrating the two levels, the model can be used for many tasks, such as online clustering of trajectories, anomaly detection, or reasoning about future movements and goal locations. These high level understandings about human motion patterns and behaviors are crucial for many robotics applications. For self-driving cars, the model provides predictions of how and where pedestrians will walk around cars, roads and intersections \cite{hardy2015multi}. For surveillance, the model gives locations for a patrol robot to intercept pedestrians. For personal robotics, the model provides predictions for collision avoidance in all types of environments (museums, at home, etc.).

\begin{figure}[!t]
  \centering
  \includegraphics[width=70mm,scale=1]{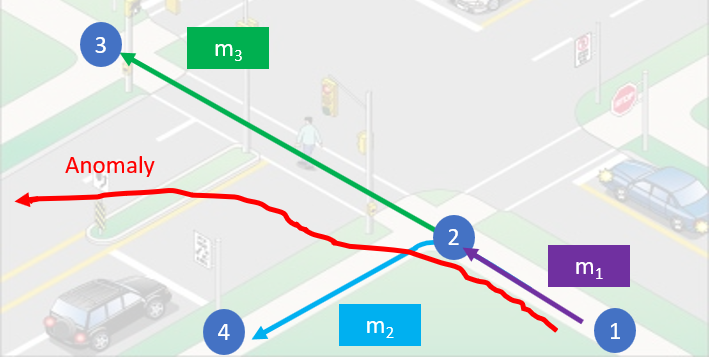}
  \caption{A motivating example for the motion model in this paper. The local motion patterns are labeled $m_{1},m_{2},m_{3}$ and shown in purple, light blue, and green respectively. There are four transition points annotated with blue bubbles. There is an anomalous trajectory shown in red. Background image from \cite{PinHawk}.}
  \vspace{-.3cm}
  \label{Fig:Highlights}
\end{figure}

Fig. \ref{Fig:Highlights} provides a motivating example for the motion model in this paper. The pedestrian movement patterns are decomposed into continuous motion patterns and discrete transition points. Transition probabilities are learned between the states, which are both motion patterns and transition points, the pedestrian can be in. The model is able to perform high level inference about pedestrian motion in the scene and also detect anomalous pedestrian behaviors.

\subsection{Related Work}

Pedestrian motion models have been developed from trajectory data in several ways. One approach involves using Bayesian non-parametric models to cluster trajectories, such as Gaussian Processes (GP) and Dirichlet-Process Gaussian Processes (DPGP) \cite{TrajModelGP}, \cite{TrajDPGP}, \cite{MotionPatternsBNP}. A learned GP over a cluster of trajectories is referred to as a motion pattern, which maps $\{x,y\}$ positions to a velocity flow field \cite{TrajDPGP}. While these models have the flexibility to fit variable trajectory data and identify clusters of trajectories, they do not capture higher level transitions, such as trajectories branching, or discrete pedestrian decisions such as stopping at an intersection. 
Although these non-parametric models capture coarse behaviors, higher level goals of transition points are important for subsequent reasoning. Other models using Bayesian non-parametrics have combined the temporal learning advantages of long short-term memory (LSTM) networks with the flexibility of GPs \cite{ForecastLSTM}. Another model uses LSTM by itself to learn movement and predict motion \cite{SocialLSTM}.  However, these models also do not capture higher level transitions corresponding to changes in pedestrian behavior, and are only meant for short term trajectory predictions.

Another approach uses discrete state transition models to learn trajectory motion, such as Hidden Markov Models (HMM) \cite{DirectionalHMM}, \cite{MiningTrajHMM}, \cite{PredFutHMM}, \cite{HMMIndoorMobile}, \cite{GHMM}. One of the challenges with using HMM is defining discrete latent states that capture high level motion characteristics. In contrast, in this paper the discrete states are intuitively defined as transition points, which are formally inferred from low level trajectory data. Some HMM approaches use a grid representation to define states. These models have trouble identifying motion patterns and higher level behaviors among trajectory clusters.  Grid based approaches also can lose valuable information when real trajectories do not fit cleanly into the grid space. Other models identify transition states as locations where many trajectories converge in a gridded space \cite{ModPredSubgoal}. In these cases, information about trajectory motion patterns is lost and complex transition points are not typically modeled. A more general approach to defining and modeling transitions between general discrete states is desired.

The work closest to this paper uses a dictionary learning algorithm to discover local motion patterns which are similar to sub-trajectory clusters \cite{ASNSC}. This model cannot learn intuitive discrete transition points, such as trajectories branching and does not consider multiple future transitions between subtrajectory clusters. There are many other approaches to modeling movement patterns such as Vector Field \textit{k}-Means \cite{VFkMeans}, finding subtrajectories by clustering line segments \cite{TrajCPandG}, and finding frequent patterns through error metric clustering \cite{FreqTrajClustMining}. These methods identify common patterns and cluster trajectories accordingly. However, high level trajectory behavior at transition points is not captured. A model that captures high level discrete decisions is required for higher level reasoning.

Generally, data-driven trajectory models from the community have also not considered anomaly detection online, which is addressed in the model presented in this paper.

This paper provides the following contributions:

\begin{itemize}
\item  A novel iterative probabilistic clustering algorithm, which finds low level clusters of sub-trajectories using DPGP and discovers high level transition points with hypothesis testing. The number of clusters and transition points is not pre-specified.
\item Online probabilistic prediction of both trajectories and novel high level transition points.
\item A formal way to discover anomalous trajectories online
\end{itemize}



\section{PEDESTRIAN MOTION MODEL}

\begin{figure}[thpb]
  \centering
  \includegraphics[width=90mm,scale=1]{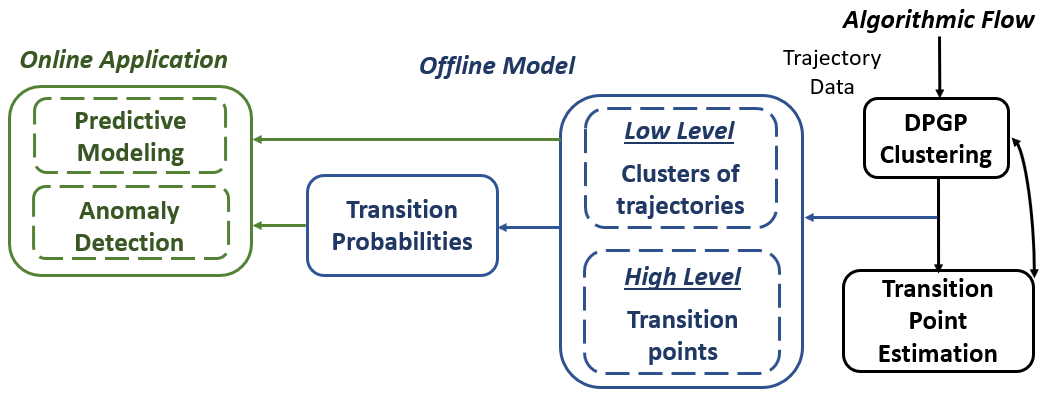}
  \caption{Algorithmic flow of iterative DPGP clustering and hypothesis test transition point estimation (black) of the proposed pedestrian model (blue) for both low level trajectories and high level transitions. The model is then used for online inference (green).}
  \vspace{-.3cm}
  \label{Fig:BlockDiagram2}
\end{figure}

Fig. \ref{Fig:BlockDiagram2} shows a block diagram of the proposed pedestrian model and algorithmic flow. First, raw trajectory data are input into an iterative algorithm with Dirichlet Process Gaussian Process (DPGP) clustering and hypothesis testing, which produces trajectories clustered into sub-trajectory motion patterns and estimated transition points. The clusters and transition points are then used to find transition probabilities. Finally, in online application, the motion patterns, transition points, and transition probabilities are used for predictive modeling and anomaly detection.

\section{MOTION \& TRANSITION POINT MODELING}

\subsection{DPGP Trajectory Clustering}

The pedestrian dataset is composed of the set of $n_{t}$ trajectories $T = \{t^{\textit{1}}\!\!,... t^{\textit{k}}\!\!,...t^{\textit{n\textsubscript{t}}}\}$, where $t^{k}$ is an individual trajectory. Each trajectory $t^{k}$ is divided into a sequence of $n^{k}_{p}$ points $[\{x^{k}_{\textit{1}},y^{k}_{\textit{1}}\},...\{x^{k}_{\textit{i}},y^{k}_{\textit{i}}\},...\{x^{k}_{{n^{k}_{p}}},y^{k}_{{n^{k}_{p}}}\}]^{T}$, $t^{k}_{i} = \{x^{k}_{i},y^{k}_{i}\}$ represents the \textit{i}\textsuperscript{th} data point, and $\{x,y\}$ represents positional data. The time interval between two consecutive points $i$ and $i+1$ in a trajectory is constant. A motion pattern is defined as a mapping from $\{x,y\}$ to $\{\dot{x},\dot{y}\}$, where $\{\dot{x},\dot{y}\}$ are the time derivatives of the $x$ and $y$ positions respectively. The motion pattern enables a distribution of velocity to be estimated given any 2-D point in space. The set of motion patterns $M$ is defined as $\{m_{\textit{1}},...m_{j},...m_{n\textsubscript{\textit{m}}}\}$, where $m_{j}$ is an individual motion pattern and $n_{m}$ is the total number of motion patterns in $M$. Each motion pattern $m_{j}$ contains $n_{m\textsubscript{\textit{j}}}$ trajectories, and there are no shared trajectories between motion patterns. The goal of DPGP trajectory clustering is then to divide the trajectory data $T$ into $n_{m}$ representative motion patterns.

The Dirichlet Process (DP) \cite{DPTeh} is a nonparametric clustering process that in this case discovers a mixture of GP models. DP allows for automatic discovery of the number of clusters which grows as the data becomes more complex. Gibbs sampling is used for the clustering process of DP. An initial number of clusters is manually chosen and the cluster assignments are randomly initialized for the DP. In the clustering process, each trajectory $t^{k}$ belongs to an existing motion pattern $m_{j}$ with prior probability

$$
\frac{n_{m\textsubscript{\textit{j}}}}{\alpha+n_{t}-1}
\eqno{(1)}
$$
and belongs to a \textit{new} motion pattern with prior probability

$$
\frac{\alpha}{\alpha+n_{t}-1}
\eqno{(2)}
$$
where $\alpha$ is a concentration parameter. 

The GP is used as a likelihood metric for the DP; GP \cite{PRMLBishop} models motion patterns by mapping the set $\{x,y\}$ to the set $\{\dot{x},\dot{y}\}$. The proposed GP uses a radial basis function (RBF) kernel \cite{TrajDPGP} to weight local data, defined as
	
$$
K_{x}(x,y,x',y')=\sigma_{x}^2\exp(-\frac{(x-x')^2}{2u_{x}^2}-\frac{(y-y')^2}{2u_{y}^2})+
$$
$$	
\sigma_{n}^2\delta(x,y,x',y') 
\eqno{(3)}
$$
where $K_{x}$ is the kernel function in the $x$ direction, $\sigma_{x}$ and $\sigma_{n}$ are standard deviations for the $x$ direction and noise respectively, and $u_{x}$ and $u_{y}$ are length scales in the $x$ and $y$ directions respectively. $\delta$ is the dirac delta function. A separate kernel $K_{y}$ is computed for the $y$ direction and is defined similarly. The kernel function represents the covariance between the set of points $\{x,y\}$ and $\{x',y'\}$.

The predicted velocity of a motion pattern at a candidate new point $\{x^{*},y^{*}\}$ is $f^{*} = \{\dot{x^{*}},\dot{y^{*}}\}$ with the distribution
$$
f^*\sim \mathcal{N}(K_{*}^{T}K^{-1}f,K_{**}-K_{*}^{T}K^{-1}K_{*})  \eqno{(4)}
$$
where $K=\{K_{x}, K_{y}\}$, $K_{*}=K(x,y,x_{*},y_{*})$, $K_{**}=K(x_{*},y_{*},x_{*},y_{*})$, and $f$ represents the training data \{$\dot{x}$,$\dot{y}$\} from the motion pattern. 

The GP likelihood of a trajectory belonging to a motion pattern $m_{j}$ is chosen as
$$
l(t^{k}, m_{k}=m_{j}|\theta)=\prod_{i=1}^{n^{k}_{p}}p(\dot{x}_{i}^{k}|t^{k},m_{j},\theta) \cdot
\eqno{(5)}
$$

$$
\prod_{i=1}^{n^{k}_{p}}p(\dot{y}_{i}^{k}|t^{k},m_{j},\theta)
$$
where $\theta$ are the GP hyperparameters of $m_{j}$, and $\dot{x}_{i}^{k}$ and $\dot{y}_{i}^{k}$ correspond to the \textit{i}\textsuperscript{th} point in $t^{k}$. 

An issue with using GPs as a likelihood metric is the planar shift problem \cite{MotionPatternsBNP}. In \cite{MotionPatternsBNP} a grid based approach is used instead of GPs as a likelihood metric. However, a grid based approach loses the flexibility of GPs, so in this paper the likelihood is adjusted to address the planar shift problem. The GP only accounts for the differences in velocity distribution in (5), so a weighting term to account for the positional distribution of trajectories is added into the likelihood
$$
l_{w}(t^{k}, m_{k}=m_{j}|\theta_{j})=l(t^{k}, m_{k}=m_{j}|\theta_{j}) \cdot
\eqno{(6)}
$$

$$
\prod_{i=1}^{n^{k}_{p}} w(x_{i}^{k},y_{i}^{k},m_{j},\epsilon,\beta)
$$

$$
w(x_{i}^{k},y_{i}^{k},m_{j},\epsilon,\beta)=\bigg(1+\frac{n_{m\textsubscript{\textit{j,$\epsilon$}}}}{n_{m\textsubscript{\textit{j}}}}\bigg)^{\beta}
\eqno{(7)}
$$
where $\epsilon$ is a parameter for a neighborhood around $\{x_{i}^{k},y_{i}^{k}\}$, $\beta$ is a weighting parameter, $n_{m\textsubscript{\textit{j,$\epsilon$}}}$ is the number of trajectories in $m_{j}$ that pass through the $\epsilon$ neighborhood surrounding $\{x_{i}^{k},y_{i}^{k}\}$, and $w(x_{i}^{k},y_{i}^{k},m_{j},\epsilon,\beta)$ is a weighting term that accounts for the positional distribution of $t^{k}$ with respect to $m_{j}$.

\begin{figure}[thpb]
  \centering
  \includegraphics[width=60mm,scale=1]{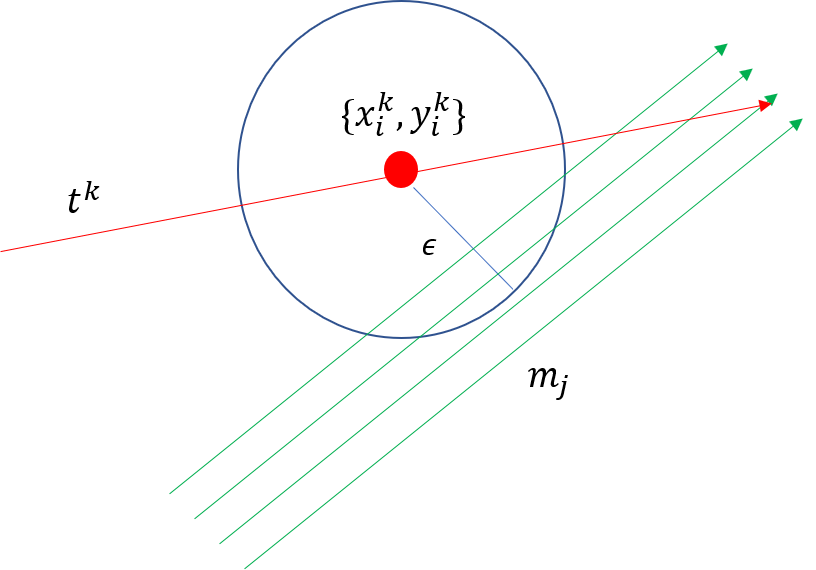}
  \vspace{-.2cm}
  \caption{The $\epsilon$ neighborhood of point $\{x_{i}^{k},y_{i}^{k}\}$. The $\epsilon$ neighborhood is in blue, $m_{j}$ is in green, $\{x_{i}^{k},y_{i}^{k}\}$ is the red dot, and $t^{k}$ is in red.
  }
  \vspace{-.3cm}
  \label{Fig:Neps}
\end{figure}



The positional weighting term $w$ intuitively accounts for the percent of trajectories of $m_{j}$ in the space near each point in $t_{k}$, and weighs if the positional distribution of $t^{k}$ matches the positional distribution of trajectories in $m_{j}$.  The parameter $\beta$ is a tuning parameter for how much the positional distribution is accounted for in the final likelihood.

The probability of a trajectory $t^{k}$ being assigned to an existing cluster $m_{j}$ in the DPGP can be defined as

$$
p(m_{k}=m_{j}|t^{k},m_{j},\theta) \propto \frac{n_{m\textsubscript{\textit{j}}}}{\alpha+n_{t}-1} \cdot
$$
$$
l_{w}(t^{k}, m_{k}=m_{j}|\theta)
\eqno{(8)}
$$
and the probability of being assigned to a new cluster is 

$$
p(m_{k}=m_{n_{m}+1}|t^{k},m_{n_{m}+1},\theta) \propto \frac{\alpha}{\alpha+n_{t}-1} \cdot
$$
$$
\int l_{w}(t^{k}, m_{k}=m_{n_{m}+1}|\theta_) d\theta
\eqno{(9)}
$$

Algorithm 1 provides pseudocode of the DPGP clustering process. In practice, the algorithm generally converges quickly within five iterations.

\begin{algorithm}[!b]
 \caption{DPGP Trajectory Clustering}
 \begin{algorithmic}[1]
 \renewcommand{\algorithmicrequire}{\textbf{Input:}}
 \renewcommand{\algorithmicensure}{\textbf{Output:}}
 \REQUIRE Trajectory dataset $T$
 \ENSURE  Trajectories clustered into motion patterns $M$
 \\ \textit{Initialization} : each trajectory $t^{k}$ is randomly grouped into one of $n_{m}$ clusters, where $n_{m}$ is manually chosen as the initial number of trajectories
 \FOR {$iterations$}
  \FOR {$i = 1$ : $n_{t}$} 
  \FOR {$j = 1$ : $n_{m}$}
  	\STATE calculate $p(m_{t^{i}}=m_{j}|t^{i},m_{j},\theta)$
  	\ENDFOR
  \STATE calculate $p(m_{t^{i}}=m_{n_{m}+1}|t^{i},m_{n_{m}+1},\theta)$
  \STATE assign $t^{i}$ to an existing cluster or start a new cluster based on the highest calculated posterior.
   \ENDFOR
   \STATE set $n_{m}=$ number of clusters
  \ENDFOR
 \RETURN 
 \end{algorithmic} 
 \end{algorithm}
  
\subsection{Transition Point Estimation}

Transition points are defined as spatial areas where motion patterns undergo transition behavior, such as discrete branching and merging or discrete changes in velocity. Transition points typically reflect discrete decisions by the pedestrian when walking, such as branching or merging at a fork in the road, or stopping at an intersection. Transition points separate sub-trajectories when appropriate. Formally, we define a transition point as a Gaussian distribution over spatial 2-D points where transition behavior occurs. 
 
Branching or merging is defined when sub-trajectories, or subsets of the clustered trajectories, are shared. Given two motion patterns, the sub-trajectory is found via a two-tailed hypothesis test. Consider a point $q$ that is part of a trajectory in motion pattern $m_{\textrm{1}}$, and a second motion pattern $m_{\textrm{2}}$. It is desired to discover if $m_{\textrm{1}}$ and $m_{\textrm{2}}$ have a shared sub-trajectory, and if $q$ is a point in the sub-trajectory. The null hypothesis $H\textsubscript{0}$ and alternative hypothesis $H\textsubscript{A}$ are defined as 
 
 $$
 H\textsubscript{0}: \textrm{\textit{q} $=$ part of a shared sub-trajectory} \eqno{(10)}
 $$
 $$
 H\textsubscript{A}: \textrm{\textit{q} $\neq$ part of a shared sub-trajectory} \eqno{(11)}
 $$
The test statistic is calculated as

 $$
t_{q} \propto g(q,f\textsubscript{\textit{q},2},f\textsubscript{\textit{q},1},\theta)w\textsubscript{$\epsilon$}(q,m\textsubscript{2},\epsilon,\beta)
 \eqno{(12)}
 $$
 $$
 g=p(q,f\textsubscript{\textit{q},2}|f\textsubscript{\textit{q},1},\theta)
 $$
where the quantity $t_{q}$ is the test-statistic for $q$. $f\textsubscript{\textit{q},1}$ is the predicted distribution of velocity at $q$ given the $m_{\textrm{1}}$ GP, and $f\textsubscript{\textit{q},2}$ is the predicted distribution of velocity at $q$ given the $m_{\textrm{2}}$ GP. The term $w\textsubscript{$\epsilon$}(q,m\textsubscript{2},\epsilon,\beta)$ is the same weighting term from (6). If the \textit{p}-value calculated from $t_{q}$ is less than the specified significance level $\sigma$, then $H\textsubscript{0}$ is rejected. In other words, given a single $\{x,y\}$ point $q$ in $m_{1}$, it is desired to find if $q$ has a matching velocity and positional distribution with $m_{2}$. For each trajectory in $m_{1}$, all points with overlapping velocity and positional distributions with $m_{2}$ are found, giving potential sub-trajectories.
 
 The branch or merge transition points are either at the beginning or end of a sub-trajectory depending on whether the trajectories are joining together or splitting apart. The points adjoining the sub-trajectories and defined to be branch or merge points are fitted with a normal distribution. 
 
\begin{figure}[thpb]
  \centering
  \includegraphics[width=80mm,scale=1]{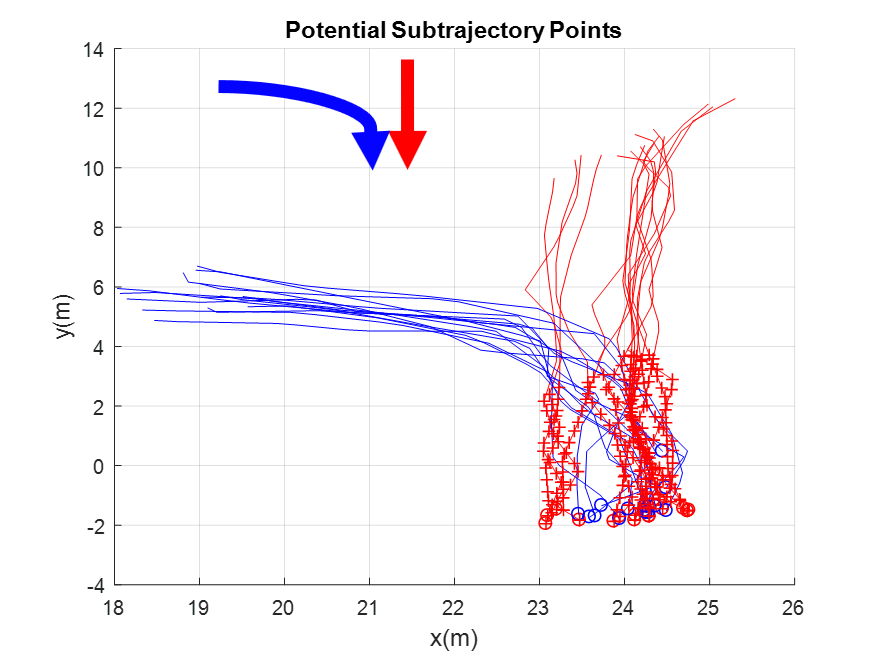}
  \vspace{-.2cm}
  \caption{Discovery of sub-trajectories for the case of merging motion patterns. There are two motion patterns in red and blue. Arrows in the top left show the directionality of the motion patterns. Circles on the trajectories indicate where the trajectory data ends. Potential sub-trajectory points in the red pattern are indicated with red crosses. The data is from video 3 of the DukeMTMC project \cite{DukeMTMC}.}
  \vspace{-.3cm}
  \label{Fig:SubtrajDuke3}
\end{figure}

Fig. \ref{Fig:SubtrajDuke3} shows potential sub-trajectory points found for merging motion patterns in the Duke MTMC dataset \cite{DukeMTMC}. The sub-trajectory points found intuitively appear in the region of overlap between the two motion patterns. The spatial distribution over the location where the red pattern merges with the blue pattern is a transition point.
 
\subsection{Iterative Clustering}

The iterative clustering process uses both DPGP clustering and sub-trajectory discovery. The algorithm starts with raw trajectory data and no cluster assignments. 
 Algorithm 2 provides the pseudocode for the iterative clustering.

\begin{algorithm}[!b]
 \caption{Iterative Trajectory Clustering}
 \begin{algorithmic}[1]
 \renewcommand{\algorithmicrequire}{\textbf{Input:}}
 \renewcommand{\algorithmicensure}{\textbf{Output:}}
 \REQUIRE Trajectory dataset $T$
 \ENSURE  Trajectories clustered into motion patterns of representative sub-trajectories $M$
  \WHILE {$true$}
  	\STATE DPGP Clustering
  	\STATE find potential sub-trajectory between motion patterns $m_{1}$ and $m_{2}$ with pairwise comparisons through hypothesis testing
  \IF{no sub-trajectories are found}
   \STATE break
  \ENDIF
  \STATE split the trajectories in the compared pair of motion patterns into smaller sub-trajectories corresponding to a new motion pattern
  \STATE $n_{m} = n_{m} + 1$
  \ENDWHILE
 \RETURN 
 \end{algorithmic} 
 \end{algorithm}

\begin{figure*}[!t]
  \centering
  \includegraphics[width=130mm,scale=1]{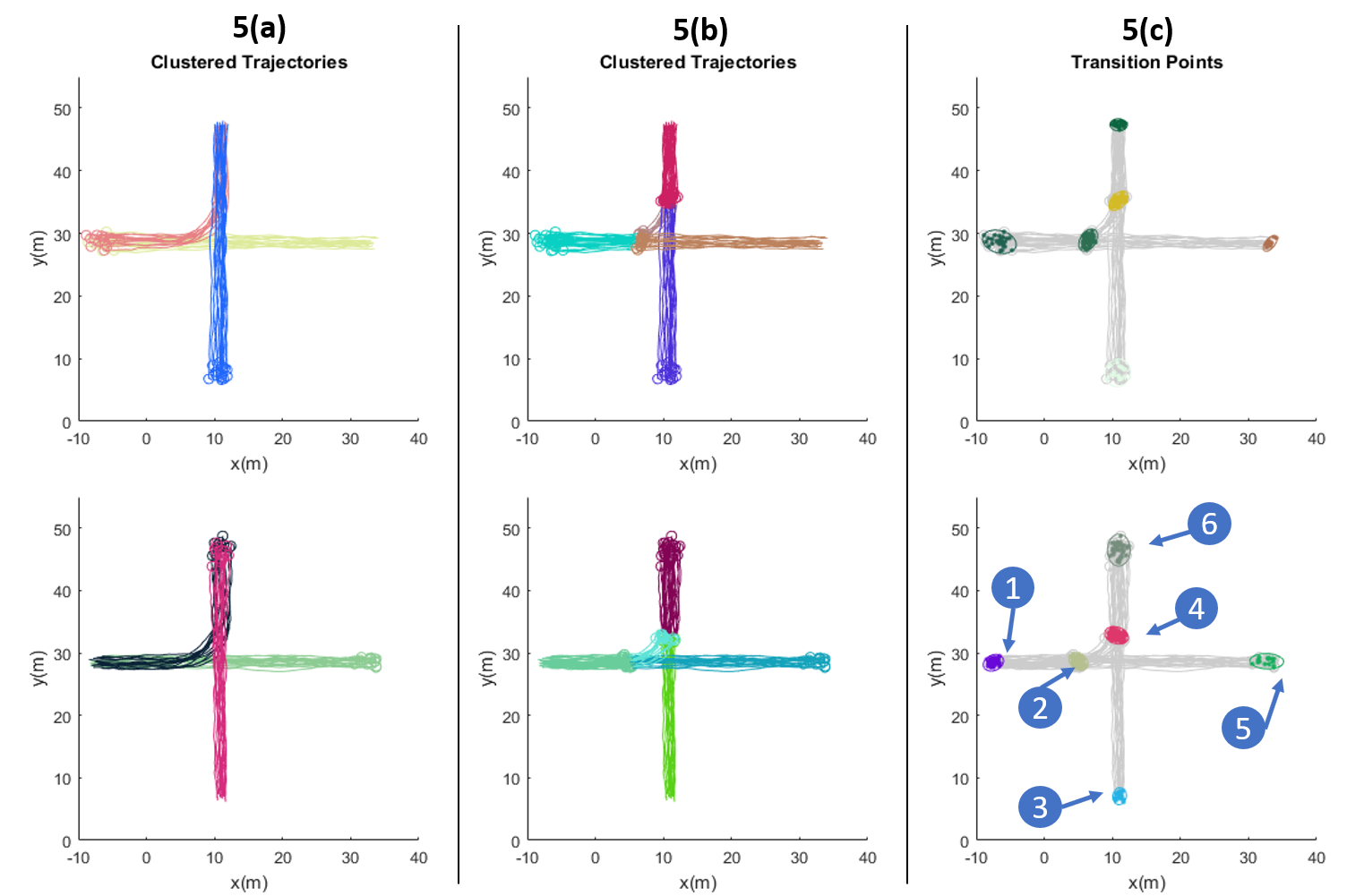}
  \vspace{-.25cm}
  \caption{Results of the iterative DPGP trajectory clustering algorithm and transition point estimation on simulated data with 90 trajectories. The trajectories move in two directions, and the top and bottom rows correspond to each of the directions respectively. \ref{Fig:ToyData2}(a) is the initial clustering result before any splitting of trajectories. \ref{Fig:ToyData2}(b) shows the iterative clustering result after convergence. \ref{Fig:ToyData2}(c) shows the transition points found with $95\%$ confidence ellipses. The blue bubbles annotating \ref{Fig:ToyData2}(c) (bottom) correspond to the states in Table I. The numbers $1-6$ in the blue bubbles correspond to the states $s_{1}-s_{6}$ respectively.}
  \vspace{-.3cm}
  \label{Fig:ToyData2}
\end{figure*}

\subsection{Transition Probabilities}
Transition probabilities are learned over the transition points and motion patterns in a transition probability matrix (TPM). Given the clustered data from the iterative DPGP clustering, DBSCAN is first used to discover the transition points. DBSCAN is a density based clustering algorithm \cite{DBSCAN}. The points at the start and ends of trajectories are inputs into DBSCAN and the output is clusters of transition points. Normal distributions are fit over the transition point clusters to find a spatial distribution.

The set of states $S$ in the TPM model are defined as $\{s_{\textit{1}},...s_{h},...s_{n\textsubscript{s}}\}$, where $s_{h}$ is either a transition point or motion pattern and $n_{s}$ is the total number of states. The TPM stores transition probabilities between the states in $S$. The probability

$$
p(s_{i}|s_{j})=\frac{d_{j}^{i}}{d_{j}}
\eqno{(13)}
$$
is the $ij^{th}$ element of the TPM, $d_{j}$ is the number of elements or trajectories in $s_{j}$, and $d_{j}^{i}$ is the number of elements in $s_{j}$ that pass through $s_{i}$ at a future time step. The matrix contains probabilities for all possible transitions, short term and long term.
 
 Fig. \ref{Fig:ToyData2} illustrates the iterative clustering results on a simulated data set with 90 trajectories. The simulated dataset represents an intersection with a total of six trajectory clusters that can be further divided into ten sub-trajectory motion patterns and twelve transition points. The trajectories move along the paths in two directions, which are processed simultaneously in the model. 
 The iterative clustering algorithm successfully finds all ten motion patterns and all twelve transition points after iteratively splitting the initial six patterns into sub-trajectory patterns. The patterns and transition points match what intuition of the motion in the scene should be.
 
 Table I shows the TPM entries for the pedestrian states $s_{1}-s_{6}$, which are automatically extracted from the simulated data shown in Fig. \ref{Fig:ToyData2}. For brevity, Table I only includes transition probabilities for the transition points shown in Fig. \ref{Fig:ToyData2}(c) (bottom). The full TPM includes transitions probabilities between all motion patterns and transition points. The values in Table I match intuition of what transition probabilities should be between the states annotated in Fig. \ref{Fig:ToyData2}(c) (bottom).

\begin{table}[]
\centering
\caption{Transition Probabilities Between Transition Points}
\small
\begin{tabular}{|c|c|c|c|c|c|c|}
\hline
\textbf{States} & \textbf{$s_{1}$} & \textbf{$s_{2}$} & \textbf{$s_{3}$} & \textbf{$s_{4}$}       & \textbf{$s_{5}$}       & \textbf{$s_{6}$}       \\ \hline
\textbf{$s_{1}$}     & 0           & 1           & 0           & 0.43 & 0.57 & 0.43 \\ \hline
\textbf{$s_{2}$}     & 0           & 0           & 0           & 0.45 & 0.55 & 0.45 \\ \hline
\textbf{$s_{3}$}     & 0           & 0           & 0           & 1                 & 0                 & 0.93 \\ \hline
\textbf{$s_{4}$}     & 0           & 0           & 0           & 0                 & 0                 & 1                 \\ \hline
\textbf{$s_{5}$}     & 0           & 0           & 0           & 0                 & 0                 & 0                 \\ \hline
\textbf{$s_{6}$}     & 0           & 0           & 0           & 0                 & 0                 & 0                 \\ \hline
\end{tabular}
\vspace{-.5cm}
\end{table}

\section{Online Prediction of Motion and Anomaly Detection}

Once the offline model has been trained, the model can be used for (1) online predictive modeling and (2) online anomaly detection.

\subsection{Predictive Modeling}

Online predictive modeling makes use of both low level trajectory motion patterns and high level transition points. Given a new observed trajectory $t^{o}$, the algorithm attempts to cluster that trajectory with an existing motion pattern. The model assigns $t^{o}$ to the motion pattern $m_{o}$ with the highest likelihood from (6). That is, $t^{o}$ is assigned to $m_{o}$ satisfying

$$
m_{o}=\operatorname{arg\,max}_{m\textsubscript{\textit{j}}} l_{w}
\eqno{(14)}
$$
where $l_{w}$ is the likelihood from (6). Because it is desired to cluster $t^{o}$ with local sub-trajectories, a sliding window is used in online clustering with window size $w_{s}$ (in time steps of the observed trajectory data). As $t^{o}$ is observed, the last $w_{s}$ points within the trajectory are assigned to a learned model motion pattern. Given the assigned motion pattern, the model assigns probabilities to future states $S$ in the TPM.

Aside from the motion pattern assignment, $t^{o}$ may also pass through the spatial distribution of a transition point. If this occurs, the online model continues clustering with the sliding window but gives precedence to the predictive distribution of future states of the transition point.

\subsection{Anomaly Detection}

The same sliding window approach for predictive modeling is also used for anomaly detection. The maximum likelihood in (14) is compared to a predetermined threshold $l_\textrm{thresh}$ to classify a sliding window trajectory segment as anomalous. First, (14) finds $m_{o}$ which maximizes $l_{w}$. Then, if $l_{w} > l_\textrm{thresh}$ the cluster assignment $m_{o}$ is kept for predictive modeling. If $l_{w} \leq l_\textrm{thresh}$ the sliding window segment is classified as an anomaly with no predictive states.

\section{VALIDATION OF THE MOTION MODEL}
For validation, the simulated dataset shown in Fig. \ref{Fig:ToyData2} and the Duke MTMC dataset \cite{DukeMTMC} are used. The Duke dataset contains eight annotated videos taken at 60 fps on the Duke University campus. Annotations include manually labeled trajectories of pedestrians moving around a scene. Given the walking speed of the pedestrians, the data is down-sampled to 2 fps without losing any clarity in trajectories or motion patterns. The $\{x,y\}$ measurements defined in a global frame given by the Duke dataset are used as inputs to the model. 

In section V.A, the robustness of the learned model to anomalous data is evaluated using the simulated dataset in Fig. \ref{Fig:ToyData2} due to the ability to control the parameters of the simulation. The accuracy of online prediction and the anomaly detection accuracy are evaluated using the Duke dataset, given the challenges of an unstructured environment and not-ideal sensors, in sections V.B and V.C respectively.

\subsection{Robustness of the Iterative Clustering Algorithm to Anomalous Training Data}

\begin{figure}[!t]
  \centering
  \includegraphics[width=70mm,scale=1]{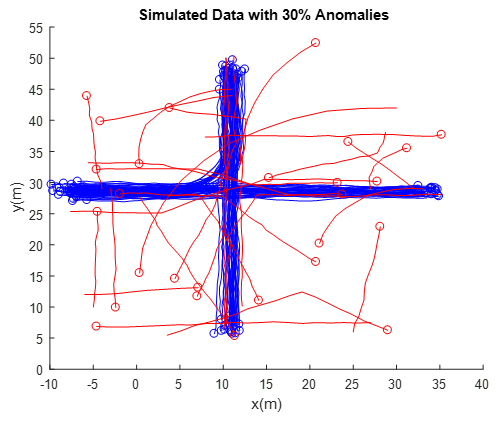}
  \vspace{-.3cm}
  \caption{Simulated dataset of 90 trajectories with $30\%$ added anomalies. The original trajectories are in blue and the anomalous trajectories are in red. The circles signify the ends of trajectories.}
  \vspace{-.4cm}
  \label{Fig:ToyData20anom}
\end{figure}

To evaluate the robustness of the learned model, varying amounts of anomalous trajectories are added to the simulated dataset (see Fig. \ref{Fig:ToyData20anom}). Anomalous trajectories are similar to adding noise to the data for learning. The amounts of anomalous trajectories added are in increments of $10\%$ of the size of the original dataset, until up to $30\%$. The resulting trajectory cluster assignments are compared to the ground truth to find the percent of trajectories that have incorrect cluster assignments. The results were averaged over 10 trials for each increment of noisy data. Fig. \ref{Fig:ToyData20anom} shows the original simulated dataset of 90 trajectories with $30\%$ anomalies added, totaling 117 trajectories.

\begin{figure}[!b]
  \centering
  \includegraphics[width=70mm,scale=1]{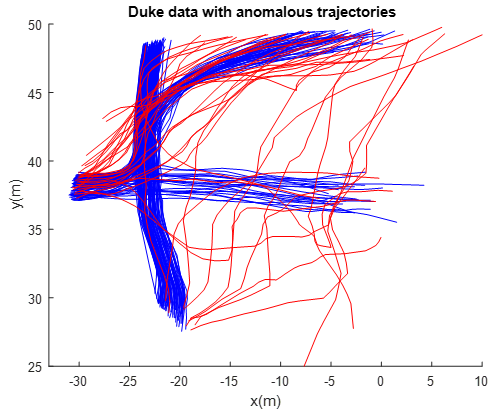}
  \vspace{-.3cm}
  \caption{Trajectories used from the Duke MTMC dataset, video number 8. The 191 trajectories used to train the offline model in k-folds validation are shown in blue with circles signifying the ends of trajectories. 40 anomalous trajectories used for online anomaly detection analysis are shown in red.}
  \label{Fig:Duke8_191}
  \vspace{-.3cm}
\end{figure}

\begin{figure*}[!t]
  \centering
  \includegraphics[width=140mm,scale=1]{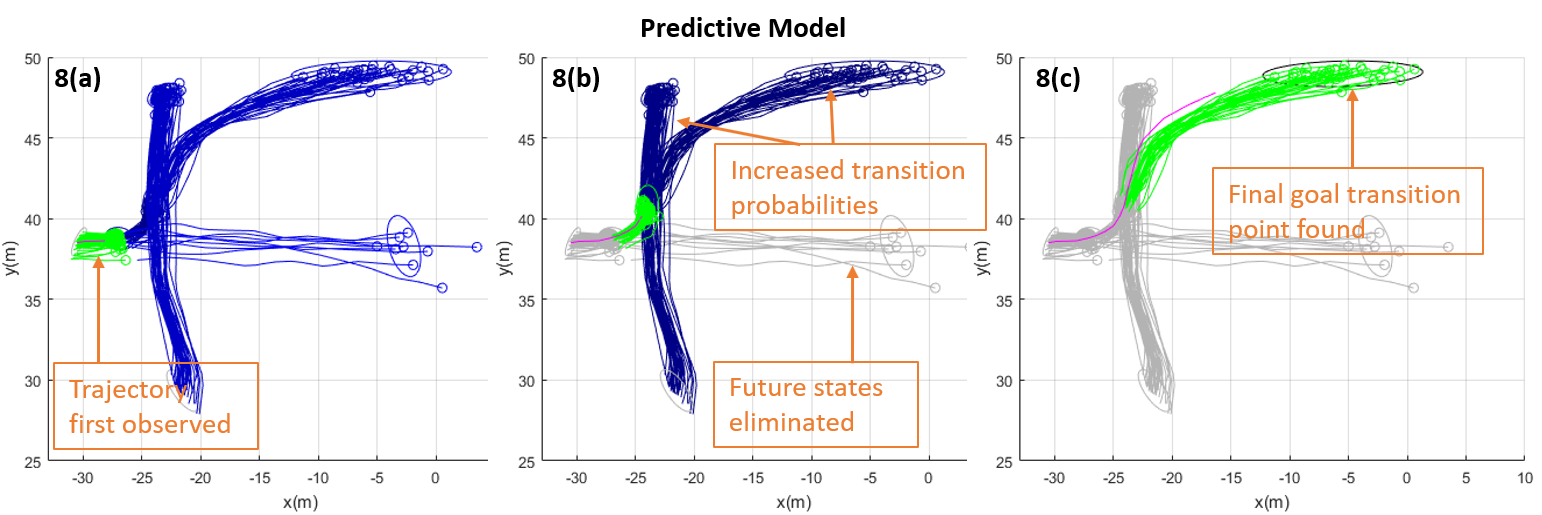}
  \caption{An example of the online predictive model. The observed trajectory is shown in magenta. The current state, which includes both transition points and motion patterns, is in green. Possible future states are in shades of blue, scaled by probability, with darker shades corresponding to higher probability of transition, and improbable future states are shaded in gray. In \ref{Fig:PredMod}(a), the trajectory has just started out so there are a wide range of possible future states. In \ref{Fig:PredMod}(b), the trajectory moves within the overlap of clusters, and the probability of transitioning to some states increases with a darker shade of blue, while other states are eliminated. In \ref{Fig:PredMod}(c) the trajectory has only one possible future state to transition to, which is the confidence ellipse shaded in black, the color corresponding to the highest transition probability of one.}
  \vspace{-.3cm}
  \label{Fig:PredMod}
\end{figure*}

Table II shows the iterative clustering algorithm is robust to anomalies with an error of $9.1\%$ with a training dataset of $30\%$ anomalies when compared with the ground truth. The ground truth takes labels from the fully trained model in Fig. \ref{Fig:ToyData2} and includes additional labels for anomalies. The algorithm is still able to extract the primary underlying motion patterns in Fig. \ref{Fig:ToyData2} and iteratively split the trajectory data accordingly, while classifying the majority of the anomalies into their own singleton clusters. The number of transition points found did not change with the added anomalies. The presence of outliers did not have a significant effect on the underlying distribution of motion patterns discovered by the algorithm.

\begin{table}[H]
\centering
\caption{$\%$ Error for Different Amounts of Anomalies}
\small
\begin{tabular}{|c|c|c|c|}
\hline
\textbf{Anomalies (\%)} & \textbf{10} & \textbf{20} & \textbf{30} \\ \hline
\textbf{Error (\%)}     & 4.6        & 6.5         & 9.1         \\ \hline
\end{tabular}
\vspace{-.2cm}
\end{table}

\subsection{Pedestrian Prediction Accuracy}

For analysis of pedestrian prediction accuracy, k-fold cross validation with $k=10$ was used. The model was evaluated on 191 trajectories extracted from the Duke MTMC dataset, video number 8. For each fold, $90\%$ of the 191 trajectories are used for training the model and the remaining $10\%$ are used to simulate online tracker readings of new trajectories; the trajectories are shown in blue in Fig. \ref{Fig:Duke8_191}. To evaluate the model, two metrics are introduced. 1) Prediction Accuracy: the percent of time in the total observed trajectory that the final state the trajectory will reach is part of the distribution of predicted states and 2) Transition Prediction Time (TPT): the number of time steps after leaving a transition point needed for the model to accurately predict the next transition point. The sliding window size $w_{s}=6$ was manually chosen to balance resolution and long term characteristics. Prediction accuracy of the proposed model is compared with a constant velocity prediction model as a baseline where the average velocity of the sliding window is extrapolated to predict future locations of the pedestrian. TPT is not applicable to a constant velocity prediction model, since high level transitional description of trajectories is the proposed models novelty and not available in the competing model.


The metrics were evaluated across the 10 folds and the results are shown in Table III.
A statistical $t$-test is performed to test the significance of the difference between predictor accuracies. The proposed model performed significantly better than the constant velocity baseline ($p<0.01$), yielding the correct final state in the predicted distribution at $95\%$ of the time on average, compared with $39\%$ for constant velocity predictions. For TPT, the proposed model finds the next transition point within 1.1s on average, i.e. 2.2 time steps at 0.5s measurement intervals, which is fast considering the speed of pedestrians. The algorithm is able to forecast long term final locations of new observed trajectories and can converge on future states fast enough for online prediction.

\begin{table}[]
\centering
\caption{K-folds Cross Validation Results}
\small
\begin{tabular}{|c|c|c|c|}
\hline
\textbf{Metrics} & \textbf{\begin{tabular}[c]{@{}c@{}}Prediction \\Accuracy ($\%$): \\ Proposed \\ Model\end{tabular}} & \textbf{\begin{tabular}[c]{@{}c@{}}Prediction \\Accuracy($\%$): \\ Constant \\ Velocity\end{tabular}} & \textbf{\begin{tabular}[c]{@{}c@{}}TPT(s): \\ Proposed \\ Model\end{tabular}} \\ \hline
\textbf{Mean}    & 95                                                                              & 39                                                                                 & 1.10                                                                             \\ \hline
\textbf{Max}    & 99                                                                              & 51                                                                                 & 1.50                                                                             \\ \hline
\textbf{Min}     & 89                                                                              & 31                                                                                 & 0.85                                                                             \\ \hline
\end{tabular}
\vspace{-.2cm}
\end{table}

Fig. \ref{Fig:PredMod} shows an example of the online predictive model when observing a new trajectory; the model is shown in one-directional flow of trajectories for brevity. Note the trajectory patterns all move in two directions. 
In Fig. \ref{Fig:PredMod}(a), the newly observed trajectory can potentially transition to most future states. In Fig. \ref{Fig:PredMod}(b), as the person moves within the first transition point, the transition probabilities toward the remaining future states are increased, and such future states are shown in a darker shade of blue than in Fig. \ref{Fig:PredMod}(a), while other improbable states are eliminated. In Fig. \ref{Fig:PredMod}(c), the observed trajectory can only transition to one future state, which includes a transition point where trajectories leave the scene. The model results match intuition and accurately predict the distribution of future states of the observed trajectory.

\begin{figure}[!t]
  \centering
  \includegraphics[width=80mm,scale=1]{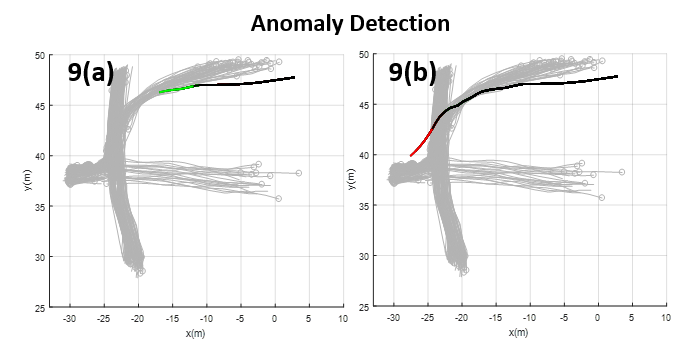}
  \caption{Online anomaly detection. The trained model is shown in gray. If the trajectory is found to be aligned with a learned motion pattern, it is highlighted in green. If the sliding window is detected as an anomaly, it is highlighted in red. The segment of the observed trajectory that is no longer in the sliding window is highlighted in black.}
  \vspace{-.6cm}
  \label{Fig:AnomDetect}
\end{figure}

\subsection{Online Anomaly Detection}


Each of the 10 folds from the split k-folds data is evaluated on accuracy of detecting anomalous trajectories online. The same 40 anomalies, which are separate from the split k-folds data, shown in red in Fig. \ref{Fig:Duke8_191} are evaluated on each training fold. The trained model is evaluated based on the ability to detect sections of trajectories that exhibit anomalous behavior based on human labels of anomalies. 

The average correct anomaly detection rate was very high at $95\%$, with a standard deviation of $3\%$. The accuracy is measured based on the percentage of anomalous trajectories where \textit{all} anomalous behavior is correctly identified by the model. Fig. \ref{Fig:AnomDetect} shows a visualization of the online anomaly detection. Fig. \ref{Fig:AnomDetect}(a) shows an example where the trajectory is initially assigned to a motion pattern. As the trajectory moves forward and additional info is collected (Fig. \ref{Fig:AnomDetect}(b)), the trajectory is labeled an anomaly once it diverges from learned motion patterns. In summary, the online anomaly detection is able to both detect (1) whether a trajectory is an anomaly and (2) which parts of the trajectory are anomalous with high accuracy.

\section{Model Complexity}

The most computationally expensive part of the model is trajectory clustering with the GPs which is $O(N^{3})$ due to inversion of matrices. Cholesky decomposition is used to speed up inversion and reduces complexity to $O(\frac{N^{3}}{3})$. The algorithm processes each new sliding window online in an average of 0.63s and is feasible to implement online. Experiments are done in MATLAB on a machine with the specifications: Intel(R) Xeon (R) CPU E5-1630 v3 @ 3.70GHz, 3701 Mhz, 4 Cores, 8 Logical Processors, 40 GB RAM, 64-bit Windows OS.

\section{CONCLUSIONS}
A probabilistic, two level pedestrian model is developed that captures low level motion patterns using an iterative Dirichlet Process Gaussian Process clustering model, high level transition points through hypothesis testing, and transition probabilities between pedestrian states. Results show that the model captures high level discrete behavior such as discrete pedestrian decisions at an intersection, while also retaining low level clusters of trajectories. The iterative model is able to learn both a high level pedestrian decision model in conjunction with low level continuous motion patterns. That is, high level transition point estimation is influenced by the low level motion pattern clustering, and vice versa. Given a new observed trajectory, the model can both quickly generate a distribution over future states and also detect anomalous behavior. The novel motion model is a potentially powerful tool for many applications, such as mobile robot path planning to intercept a person of interest at a future time or identifying anomalous pedestrian behavior at an intersection for autonomous driving.

%



\bibliography{BibDatabase}
\bibliographystyle{IEEEtran}

\end{document}